\def\year{2022}\relax
\documentclass[letterpaper]{article} % DO NOT CHANGE THIS
\usepackage{aaai22}  % DO NOT CHANGE THIS
\usepackage{times}  % DO NOT CHANGE THIS
\usepackage{helvet}  % DO NOT CHANGE THIS
\usepackage{courier}  % DO NOT CHANGE THIS
\usepackage[hyphens]{url}  % DO NOT CHANGE THIS
\usepackage{graphicx} % DO NOT CHANGE THIS
\usepackage{natbib}  % DO NOT CHANGE THIS AND DO NOT ADD ANY OPTIONS TO IT
\usepackage{caption} % DO NOT CHANGE THIS AND DO NOT ADD ANY OPTIONS TO IT
\DeclareCaptionStyle{ruled}{labelfont=normalfont,labelsep=colon,strut=off} % DO NOT CHANGE THIS
\usepackage{algorithm}
\usepackage{newfloat}
\usepackage{listings}
\floatstyle{ruled}
\newfloat{listing}{tb}{lst}{}
\floatname{listing}{Listing}
\title{Multiple-Source Domain Adaptation via Coordinated Domain Encoders and Paired Classifiers}
\author {
    Payam Karisani
}
\affiliations{
    Emory University\\
    payam.karisani@emory.edu
}
\usepackage{soul}
\usepackage{url}
\usepackage[utf8]{inputenc}
\usepackage{caption}
\usepackage{graphicx}
\usepackage{amsmath,amsfonts,amssymb}
\usepackage{booktabs}
\usepackage{subcaption}
\usepackage{mathtools}
\usepackage{makecell}

\usepackage{tabu}

\usepackage{enumitem}
\usepackage{siunitx}
\usepackage{multirow}
\usepackage{setspace}
\usepackage{algcompatible}
\usepackage[noend]{algpseudocode}
\makeatletter
\def\BState{\State\hskip-\ALG@thistlm}
\makeatother

\algdef{SE}[SUBALG]{Indent}{EndIndent}{}{\algorithmicend\ }%
\algtext*{Indent}
\algtext*{EndIndent}

\usepackage{xcolor}
\newcommand{\parm}{\mathord{\color{black!33}\bullet}}

\usepackage[switch]{lineno}  
\newcommand\METHOD{CEPC~}
\newcommand\ILLNESSDT{Illness~}
\newcommand\INCIDENTDT{Incident~}
\newcommand\VALIDATIONDT{Tuning~}

\begin{document}

\maketitle

\begin{abstract}

We present a novel multiple-source unsupervised model for text classification under domain shift. Our model exploits the update rates in document representations to dynamically integrate domain encoders. It also employs a probabilistic heuristic to infer the error rate in the target domain in order to pair source classifiers. Our heuristic exploits data transformation cost and the classifier accuracy in the target feature space. We have used real world scenarios of Domain Adaptation to evaluate the efficacy of our algorithm. We also used pretrained multi-layer transformers as the document encoder in the experiments to demonstrate whether the improvement achieved by domain adaptation models can be delivered by out-of-the-box language model pretraining. The experiments testify that our model is the top performing approach in this setting.

\end{abstract}

%%%%%%%%%%%%%%%%%%%%%%%%%%%%%%%%%%%%%%%%%%%%%%%%%%%%%%%%%%%%%%%%%%%%%%%%
\section{Introduction} \label{sec:intro}
Modern classifiers typically rely on large amount of training data. Collecting large training sets is expensive and in some cases very challenging, e.g., in legal domain \cite{legal-domain} or in social media domain \cite{view-distill,our-corona}. There exist several techniques to address the lack of training data, one of which is Domain Adaptation \cite{dom-ada-first}, where a classifier is trained in one domain (the \textit{source} domain) and evaluated in another domain (the \textit{target} domain). One of the fundamental assumptions of text classification is that training and test data follow an identical distribution. A classifier trained in one domain, typically is a poor predictor for another domain \cite{dom-ada-theory}. Therefore, Domain Adaptation primarily tackles the domain shift challenge \cite{domain-shift}. One of the major areas of Domain Adaptation, which is also the subject of our study, is the unsupervised setting where there is no labeled target data available. 

% The real world scenarios of such a case include online monitoring systems where existing domains continuously evolve and the system must adapt to newly added domains, e.g., in an online store.

Theoretical studies \cite{dom-ada-theory} have shown that the outcome of Domain Adaptation is primarily influenced by the performance of the classifier in the source domain and the discrepancy between the source and the target domains. The former can be improved irrespective of domain shift, therefore, existing domain adaptation models primarily focus on the latter. These models approach the task either explicitly by minimizing the divergence between the two distributions \cite{dan} or via a binary classifier to increase the domain confusion and to reduce the divergence \cite{gradient-reversal}. In either case, the aim is to obtain domain-invariant features between the source and the target \cite{how-transferable}.

% \footnote{We have adopted the standard Domain Adaptation terminology to describe our model and common practices to carry out the experiments. Please see the studies by \citet{transfer-survey} and \citet{sentiment-da} for more information in this regard.}
In this article, we study multiple-source Domain Adaptation \cite{dom-ada-multi}. In this setting, there are multiple labeled source domains available and the goal is to minimize the classification error in the target domain. While the availability of multiple sources provides us with more opportunities, this also comes with challenges. Naively applying single-source domain adaptation models in a multiple-source setting may cause \textit{negative transfer} \cite{transfer-survey}, i.e., deterioration in performance. In this work, we propose an objective function to minimize negative transfer via the coordination between the representations of source domains. Additionally, based on the intuition that every source classifier may specialize in predicting some regions of the target domain, we present a heuristic criterion to guide the weak source classifiers by the reliable ones.

We evaluate our model in two datasets consisting of user-generated data in Twitter. We select this domain because it is considerably less explored compared to other domains, e.g., sentiment analysis. The documents in this domain are short and their language is highly informal \cite{co-decomp}. Additionally, due to the nature of the tasks in this domain the class distributions are typically imbalanced \cite{wespad}. Taken together, these are significant challenges for existing methods. The results in this setting testify that our model is the most robust approach across various cases and consistently outperforms multiple state of the art baselines.

% Furthermore, in order to investigate whether the results achieved by our model (and existing domain adaptation baselines) can be offered by out-of-the-box language model pretraining, we use pretrained multi-layer transformers as the document encoder in all of the models used in the experiments. The results in the real world scenarios of multiple-source Domain Adaptation indicate that in certain cases existing models fail to yield any improvement over pretraining. However, the results testify that our model is the most robust approach across various cases and consistently outperforms exiting baselines.

%%%%%%%%%%%%%%%%%%%%%%%%%%%%%%%%%%%%%%%%%%%%%%%%%%%%%%%%%%%%%%%%%%%%%%%%
\section{Related Work} \label{sec:rel-work}

\citet{dom-ada-multi} theoretically show that, in multiple-source domain adaptation setting, simply combining the source classifiers yields a high prediction error. They show that there always exists a weighted average of the source classifiers that has a low error prediction in the target domain--their weights are derived from the source distributions. Therefore, existing multiple-source models typically rely on either inventively aggregating source data \cite{da-mixture} or crafting source classifiers \cite{cocktail}. While studies occasionally make an attempt to theoretically justify their approach \cite{dan}, majority of existing domain adaptation studies rely on intuitions and heuristics \cite{adda,dom-ada-distil,ccl}. In this work, we show the same pragmatism and propose an intuitive objective function to reduce the risk of negative transfer, and also present a probabilistic heuristic to encourage the collaboration between source classifiers.

Our model, which we call Coordinated Encoders and Paired Classifiers (\METHOD\!), possesses two distinct qualities. First, it employs the intuitive idea of coordination between source encoders. The aim of coordination is to organize source domains such that the domains with similar characteristics share the same encoder. Existing studies resort to different approaches. For instance, \citet{moment-match} share one encoder between all source domains, however, aim to reduce the discrepancy between them via a new loss term. \citet{da-bandit} incorporate multiple discrepancy terms and dynamically select source domains. \citet{dom-ada-curriculum-2} dynamically select data points and \citet{dom-ada-mixture-bert} dynamically aggregate source classifiers.

The second quality of our model is to exploit the properties of source data to pair source classifiers via a heuristic criterion. Here, we aim to guide the weak classifiers by the more reliable ones, therefore, the weak classifiers can adjust their class boundaries based on the information that is not available in their training data. Previous studies have explored the interaction between source classifiers using different approaches. \citet{dom-ada-distil} adversarially train source classifiers and use the distance between source and target domains to fine-tune their model. Different from their work, in our model source classifiers are paired and can interact. \citet{ccl} employ the KL divergence term as an objective term, however, their model is developed for the multi-target setting. Other relevant approaches in this area include combining semi-supervised learning and unsupervised domain adaptation to guide source classifiers \cite{semi-sup-dom-ada-cotrain}, and using attention mechanism to assign target data points to source domains \cite{dom-ada-multi-attention}. 

% In summary, to our knowledge, \METHOD is the first model that employs an objective function to coordinate source encoders and alleviate negative transfer. It is also the first model to use a novel criterion to pair source classifiers based on their prediction risk in the target domain. Below, we present our model.

%%%%%%%%%%%%%%%%%%%%%%%%%%%%%%%%%%%%%%%%%%%%%%%%%%%%%%%%%%%%%%%%%%%%%%%%
\section{Proposed Method} \label{sec:method}

In multiple-source unsupervised domain adaptation, we are given a set of $M$ source domains $S_1, S_2, \dotsc, S_M$ denoted by ${\{(\textsc{x}_{i}^{s_j},y_{i}^{s_j})\}}_{i=1}^{n_{s_j}}$ where $(\textsc{x}_{i}^{s_{\parm}}, y_{i}^{s_{\parm}})$ is the \textit{i-th} labeled document in $S_{\parm}$ and $n_{s_{\parm}}$ is the number of documents in $S_{\parm}$. We are also given a target domain $T$ denoted by ${\{\textsc{x}_{i}^{t}\}}_{i=1}^{n_t}$, where $\textsc{x}_{i}^{t}$ is the \textit{i-th} unlabeled document and $n_t$ is the number of documents in the domain $T$. The aim is to train a model which minimizes the prediction error in the target domain using the labeled source data. Note that the distributions of the documents in the source domains and the target domain are different, thus, a classifier trained on the source domains is typically a poor predictor for the target domain. 

\begin{figure} %[!ht]
\centering
\includegraphics[width=2.8in]{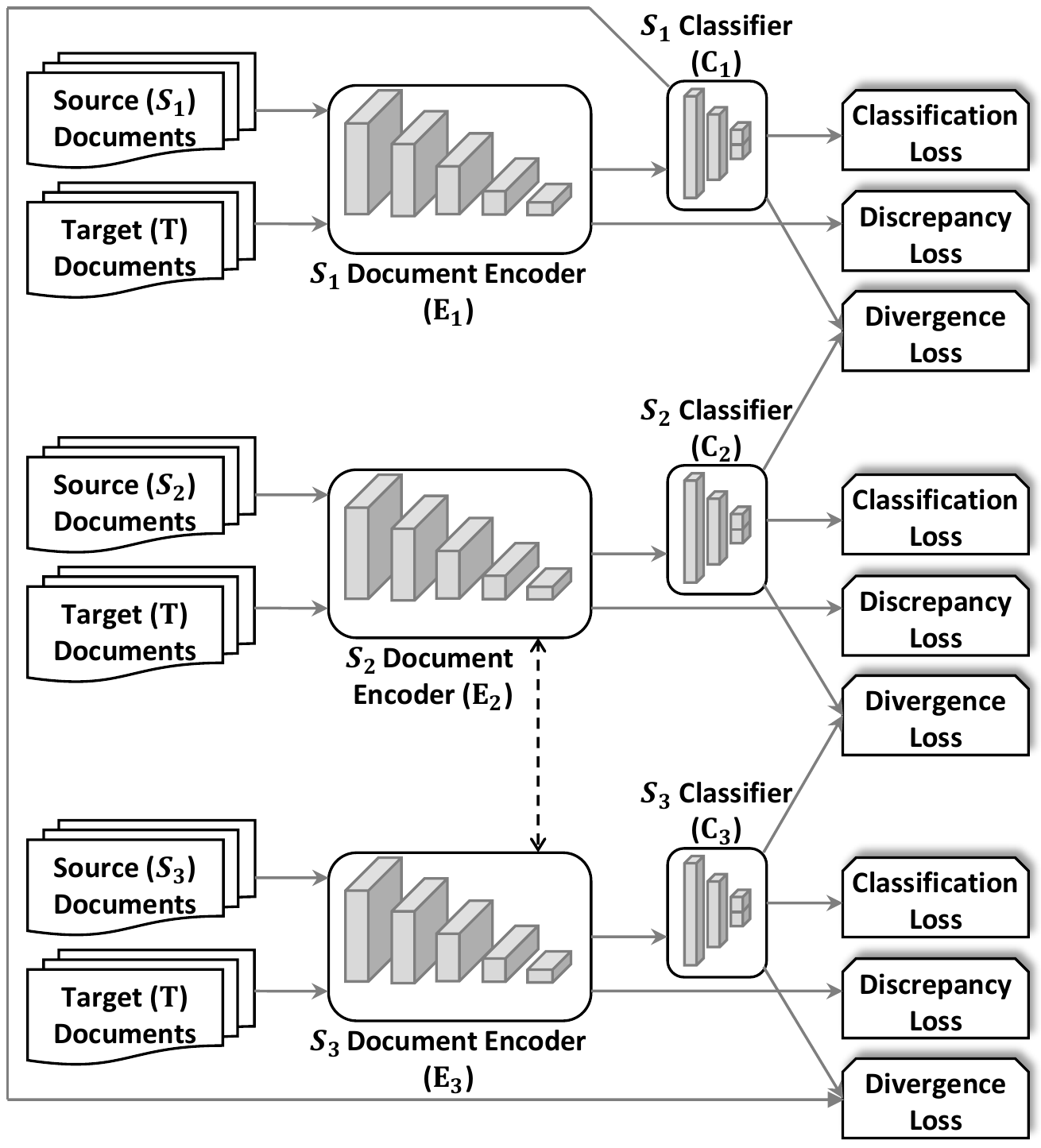}
\caption{Overview of our model for a task with three source domains. The output of each domain encoder is used in the corresponding source classifier. To train each pair of (encoder, classifier) three loss terms are minimized: 1) the regular classification loss, 2) the discrepancy loss between the encoded source and target data (see Section \ref{subsec:train}), and 3) the divergence loss between the output of the source classifiers (see Section \ref{subsec:pairing}). The dashed line indicate shared parameters (see Section \ref{subsec:coordinate}).} \label{fig:model}
\end{figure}

Figure \ref{fig:model} illustrates our model (\METHOD\!). Here, we assume that there are three source domains available. For each source domain, there is an encoder ($E$) and a classifier ($C$). Each pair is trained by three objective terms, namely: the classification loss, the discrepancy loss, and the divergence loss. The classification loss, which is the Negative Log-Likelihood, quantifies the calibration error of the classifier in the labeled source data. The discrepancy loss, quantifies the variation between the distribution of the encoded source and target documents, and the divergence loss, which quantifies the inconsistency between the output of source classifiers. The dashed lines indicate shared parameters between the modules. At test time, the source classifiers--along their corresponding encoders--are used to label target documents, and a majority voting is used to obtain the final labels.

The encoder in our model is employed to project documents into a shared low-dimensional space. For the discussion about the classification and the discrepancy loss terms see Section \ref{subsec:train}. Here, we discuss the divergence loss, which aims to minimize the inconsistency between source classifiers. The intuition behind this loss term is that because the distribution of the data in each source domain is unique, this can potentially lead to source classifiers that specialize in particular regions of the target domain. If we succeed in detecting these regions we can exploit the information to \textit{pair} the classifiers. That is, a classifier which is expected to perform well in one region, can guide the other classifiers and adjust their class boundaries. 

Additionally, we introduce an intuitive algorithm to automatically \textit{coordinate} domain encoders. In Figure \ref{fig:model}, we see that the parameters of $E_2$ and $E_3$ are shared, while $E_1$ has an exclusive set of parameters. The common theme in the literature is to have a shared encoder between source domains \cite{moment-match,da-bandit}. While this design decision has been made for a good reason \cite{multi-task}, we argue that if negative transfer \cite{transfer-survey} occurs in the shared encoder, it can propagate in the model and can exacerbate the problem. Thus, we aim to automatically coordinate the encoders to minimize the risk of negative transfer. Below, we introduce our algorithm for coordinating encoders. Then, we propose our heuristic criterion to pair source classifiers. Finally, we explain our training procedure.

\subsection{Coordinated Domain Encoders} \label{subsec:coordinate}
To provide an effective coordination between domain encoders we base our approach on the extent in which encoders must be updated to become domain-invariant. The intuition is that if two source domains with disproportional update rates share their encoders, one domain can potentially dominate the encoder updates and leave the counterpart domain under-trained. This problem is more serious when multiple domains share one encoder and one of them introduces negative transfer. In this case, the domination can cause a total failure. In the remainder of this section, we make an attempt to formalize and implement this intuitive idea by making a connection between encoder coordination and the training procedure in single source domain adaptation models.

We begin by considering the typical objective function \cite{dan} of training a model under domain shift with a single source domain $S$ and minimizing a discrepancy term:
\begin{equation} \label{eq:single-loss}
\small
% \setlength{\jot}{0pt}
% \setlength{\abovedisplayskip}{0pt}
% \setlength{\belowdisplayskip}{0pt}
% \medmuskip=0mu
% \thinmuskip=0mu
% \thickmuskip=0mu
% \nulldelimiterspace=0pt
% \scriptspace=0pt
\begin{split}
\mathcal{L}= &\frac{1}{n_s}\sum_{i=1}^{n_s} J(\theta(\textsc{x}_{i}^{s}), y_{i}^{s}) ~ + ~ \lambda \mathcal{D}(E(\textsc{x}^{s}), E(\textsc{x}^{t})),
\end{split}
\end{equation}
where the model is parameterized by $\theta$, $J$ is the cross-entropy function, $E(\textsc{x}^{s})$ and $E(\textsc{x}^{t})$ are the output of the encoder for all source and target documents respectively, $\mathcal{D}$ is the discrepancy term, and $\lambda > 0$ is a scaling factor--the rest of the terms were defined earlier in the section. The discrepancy term $\mathcal{D}$ governs the degree in which the parameters of the encoder must update to reduce the variation between source and target representations. The examples of such a loss term include central moment discrepancy \cite{central-loss} and association loss \cite{assoc-loss}. The reason to employ the scale factor $\lambda$ is that unconditionally reducing the variation between source and target representations can deteriorate the classification performance in the source domain. This can subsequently deteriorate the performance in the target domain, which is undesirable. 

% Since it is assumed that there are no labeled target documents available, the value of $\lambda$ is usually heuristically chosen \cite{coral}.

In the case that two source domains $S_1$ and $S_2$ share one encoder, Equation \ref{eq:single-loss} is written as follows:
\begin{equation} \label{eq:double-loss}
\small
% \setlength{\jot}{0pt}
% \setlength{\abovedisplayskip}{0pt}
% \setlength{\belowdisplayskip}{0pt}
% \medmuskip=0mu
% \thinmuskip=0mu
% \thickmuskip=0mu
% \nulldelimiterspace=0pt
% \scriptspace=0pt
\begin{split}
\mathcal{L}= &\sum_{j=1}^{2} \frac{1}{n_{s_j}}\sum_{i=1}^{n_{s_j}} J(\theta(\textsc{x}_{i}^{s_j}), y_{i}^{s_j})  + 
\lambda \mathcal{F}(E(\textsc{x}^{s_1}), E(\textsc{x}^{s_2}), E(\textsc{x}^{t})),
\end{split}
\end{equation}
where $E(\textsc{x}^{s_1})$, $E(\textsc{x}^{s_2})$, and $E(\textsc{x}^{t})$ are the output of the encoder for the documents in $S_1$, $S_2$, and $T$ respectively. Here, $\mathcal{F}$ is an arbitrary discrepancy function to force the encoder to remain domain-invariant between $S_1$, $S_2$, and $T$. One example of such an objective is proposed by \citet{moment-match}. In our model, we assume this loss function is linear in terms of two discrepancy terms in the source domains, i.e.,: 
\begin{equation} \label{eq:linear-discrepancy}
\small
% \setlength{\jot}{0pt}
% \setlength{\abovedisplayskip}{3pt}
% \setlength{\belowdisplayskip}{3pt}
% \medmuskip=0mu
% \thinmuskip=0mu
% \thickmuskip=0mu
% \nulldelimiterspace=0pt
% \scriptspace=0pt
\begin{split}
\mathcal{F}=\lambda_1 \mathcal{D}(E(\textsc{x}^{s_1}), E(\textsc{x}^{t})) + \lambda_2 \mathcal{D}(E(\textsc{x}^{s_2}), E(\textsc{x}^{t})),
\end{split}
\end{equation}
where $\lambda_1$ and $\lambda_2$ are two scaling factors of the introduced linear terms. Given the generalized form of Equation \ref{eq:linear-discrepancy} for $M$ source domains--i.e., a weighted average of $M$ discrepancy terms--we coordinate source encoders based on the optimal values of $\lambda$ in each single source domain adaptation model. More specifically, we group the source domains that their corresponding optimal scale factor $\lambda$ is the same in their single source domain adaptation model. This prevents source domains from dominating the direction of the updates in the encoder by ensuring that the magnitude of the gradients of this term is roughly the same for the domains with a shared encoder. Note that this approach assumes that the primary distinction between the magnitude of the gradients of $\mathcal{D}$ are caused by the coefficients $\lambda_{\parm}$ in Equation~\ref{eq:linear-discrepancy}. While this assumption is harsh, it is justifiable by the presumption that source domains are usually semantically close and so are their representations.

Determining the optimal value of $\lambda$ is challenging because there are no target labeled documents available. One can heuristically set this value \cite{dan} or can take a subset of source domains as meta-targets \cite{da-mixture}. However, these methods ignore the characteristics of the actual target domain. Here, we use a proxy task to collectively estimate the values of these hyper-parameters using labeled source documents and unlabeled target documents. Because the obtained values are the estimations of the optimal quantities we call them the pseudo-optimal.

For a task with $M$ labeled source domains we aim to maximize the consistency between the pseudo-labels generated by the single source domain adaptation models in the target domain. Thus, we seek to find one scale factor $\lambda$ for each pair of source and target domains such that if the trained model is used to label a set of target documents, the consistency between the labels across the models is the highest. To this end, we iteratively assume each set of the pseudo-labels is the ground-truth and quantify the agreement between this set and the others. In fact, by maximizing the consistency between the pseudo-labels across the models, we validate the value of each scale-factor by the output of the other models. While there are still possibilities that all of the models at the same time converge to a bad optima and result in a \textit{unique} set of noisy pseudo-labels, it is unlikely to observe such a result because the models are neural networks and their parameter space is very large.

To formally state this idea we denote the set of $M$ scale factors by $\Lambda$. The iterative procedure mentioned above for $M$ single source models can be expressed as follows:
\begin{equation} \label{eq:correlation}
\small
% \setlength{\jot}{0pt}
% \setlength{\abovedisplayskip}{3pt}
% \setlength{\belowdisplayskip}{3pt}
% \medmuskip=0mu
% \thinmuskip=0mu
% \thickmuskip=0mu
% \nulldelimiterspace=0pt
% \scriptspace=0pt
\begin{split}
Corr(\Lambda)= &\sum_{\lambda_i \in \Lambda} \enspace \sum_{\lambda_j \in \Lambda, \lambda_i \neq \lambda_j} F1(\overline{y}_{\lambda_i}, \overline{y}_{\lambda_j}) ,
\end{split}
\end{equation}

where $Corr(\Lambda)$ is the cumulative pair-wise correlation between the set of scale factors $\Lambda$. The two sets of mentioned pseudo-labels are denoted by $\overline{y}_{\lambda_i}$ and $\overline{y}_{\lambda_j}$--more specifically, $\overline{y}_{\lambda_i}$ is the set of pseudo-labels assigned to the documents in the target domain $T$ by the classifier trained on the documents in $S_i$ and $T$ when we use $\lambda_i$ as the scale factor. In this procedure, to quantify the agreement between two labeled sets we use the F1 measure, therefore, $F1(a, b)$ is the F1 measure between the set of ground-truth labels $a$ and the predictions $b$. To obtain the best scale-factors, we select the set with the highest pair-wise correlation, i.e.,:
\begin{equation} \label{eq:pseudo-correlation}
\small
% \setlength{\jot}{0pt}
% \setlength{\abovedisplayskip}{3pt}
% \setlength{\belowdisplayskip}{3pt}
% \medmuskip=0mu
% \thinmuskip=0mu
% \thickmuskip=0mu
% \nulldelimiterspace=0pt
% \scriptspace=0pt
\begin{split}
\Lambda^*=\operatorname*{argmax}_\Lambda~~Corr(\Lambda).
\end{split}
\end{equation}

In solving the search problem above, we also add the constraint that in each single source domain adaptation problem the distribution of target pseudo-labels should be similar to that of the labels in the source domain--we measure the similarity by the Jensen-Shannon distance. This inductive bias helps us to select the scale factors that result in the target classifiers that behave similarly to the source classifiers.\footnote{For each single source model, we sort the scale factors based on the Jensen-Shannon distance between the distributions of the source labels and the target pseudo-labels. Then, we greedily iterate over the scale factors to find a set that results in an agreement rate higher than a very small threshold (0.005) from the previous step.} The procedure above results in a set of pseudo-optimal scale factors--one scale factor for each pair of source and target domains. These hyper-parameters can be used to coordinate domain encoders, i.e., to determine which source domains have a shared encoder, and to train a multiple-source unsupervised model using the generalized forms of Equations \ref{eq:double-loss} and \ref{eq:linear-discrepancy}--see Sections \ref{subsec:train} and \ref{sec:setup} for the exact choices of the loss functions. In the next section, we present our heuristic criterion to pair source classifiers and further enhance the final prediction.

\subsection{Paired Source Classifiers} \label{subsec:pairing}
Each source classifier in a multiple-source domain adaptation model is trained on a distinct set of documents. It is expected that this distinction results in a set of classifiers that may be reliable in particular regions of the target feature space and be erroneous in other regions. The intuition behind the idea of pairing source classifiers is to enhance model prediction by transferring the knowledge from the classifiers that are expected to perform well in a region to the classifiers that are expected to under-perform in that vicinity. In other words, source classifiers can potentially guide each other to adjust their class boundaries.

To formalize and implement this idea is to answer three questions: 1) how to transfer knowledge between source classifiers, 2) how to detect the regions that source classifiers are reliable in, and 3) how to incorporate these two modules into a single framework. Note that these are challenging questions; because source classifiers are trained on different sets of documents and thus, their class boundaries are potentially different. On the other hand, target data which is shared between the classifiers, is unlabeled. 

To answer the first question we exploit the pseudo-labels of target documents generated by each source classifier. More specifically, during each iteration of the training procedure we use a shared batch of target documents to be labeled by all source classifiers. Then we use these pseudo-labels to adjust the classifier outputs during the back propagation phase. To formalize this idea we use Kullback–Leibler divergence term. The KL divergence quantifies the amount of divergence between two distributions. Using this term in an objective function forces two classifiers to yield close outputs--by thwarting gradients in the direction of the reliable model (i.e., detaching it from the computational graph), we can force the weak classifier to adjust.

To answer the second question we propose a criterion (in Sections \ref{subsubsec:transform-cost} and \ref{subsubsec:classifier-capacity}) to assign each target document to a source domain. Our heuristic criterion assumes that the prediction error of a classifier under domain shift is a function of two quantities: 1) it is disproportional to the magnitude of the data transformation between source and target documents, and 2) it is proportional to the prediction capacity of the trained classifier. In other words, we aim to measure the data transformation applied to a target document, as well as the accuracy of the resulting classifier in the region that the target document is transferred to. For now we denote this indicator function by $\mathcal{I}$.

Given the two arguments above, our framework for pairing source classifiers naturally consists of the product of the KL divergence term and the indicator function $\mathcal{I}$, because one of these terms transfers the knowledge (the KL divergence term), and the other (the indicator function) governs the direction of the knowledge transfer. Thus the objective function for one source classifier must follow the loosely defined form below:
\begin{equation} \label{eq:score-form}
\small
% \setlength{\jot}{0pt}
% \setlength{\abovedisplayskip}{5pt}
% \setlength{\belowdisplayskip}{5pt}
% \medmuskip=0mu
% \thinmuskip=0mu
% \thickmuskip=0mu
% \nulldelimiterspace=0pt
% \scriptspace=0pt
\begin{split}
&\Psi(S) \propto \sum_{i=1} \mathcal{I}(S,d_i) \times KL(\parm),
\end{split}
\end{equation}
where the summation is over the entire documents in the batch, $\mathcal{I}$ is our indicator function for the source domain $S$ and the document $d_i$, and $KL(\parm)$ is the KL divergence between the classifier in the source domain $S$ and all the other source classifiers. Equation \ref{eq:domain-kl} (which follows the form above) introduces our exact objective function for one source domain ($S_i$):
\begin{equation} \label{eq:domain-kl}
\small
% \setlength{\jot}{0pt}
% \setlength{\abovedisplayskip}{5pt}
% \setlength{\belowdisplayskip}{5pt}
% \medmuskip=0mu
% \thinmuskip=0mu
% \thickmuskip=0mu
% \nulldelimiterspace=0pt
% \scriptspace=0pt
\begin{split}
&\Psi(S_i) = 
\sum_{d=1}^{B}\mathcal{I}(S_i,\textsc{x}_{d}^{t})\sum_{k=1,k \neq i}^{M}~~\sum_{j=1}^{L} C_{i}(\textsc{x}_{d}^{t},j)~\log~\frac{C_{i}(\textsc{x}_{d}^{t},j)}{C_{k}(\textsc{x}_{d}^{t},j)},
\end{split}
\end{equation}
where $B$ is the number of documents in the batch, $M$ is the number of source domains, $L$ is the number of class labels, $C_i$ is the source classifier of $S_i$, and $C_{i}(\textsc{x}_{d}^{t},j)$ is the \textit{j-th} output of $C_i$ after the softmax layer for the \textit{d-th} document in the batch. $\mathcal{I}(S_i,\textsc{x}_{d}^{t})$ is our indicator function which returns 1 if the source classifier $C_i$ is the best classifier for labeling the \textit{d-th} target document, otherwise, it returns 0. 

Equation \eqref{eq:domain-kl} reduces the degree of discrepancy between source classifiers in labeling target documents. The indicator function $\mathcal{I}$ ensures that only the best source classifiers guide the other ones. 
% As we stated before, the updating procedure in this equation is uni-directional, i.e., $C_{i}(\textsc{x}_{d}^{t},j)$ is detached from the computational graph and only $C_{k}(\textsc{x}_{d}^{t},j)$ is updated--because the aim is to update the weak classifiers only. 
Given the objective term $\Psi(S_i)$ for one source domain, the final divergence loss is obtained by the summation over all of the domains:
\begin{equation} \label{eq:divergence-loss}
\small
% \setlength{\jot}{0pt}
% \setlength{\abovedisplayskip}{5pt}
% \setlength{\belowdisplayskip}{5pt}
% \medmuskip=0mu
% \thinmuskip=0mu
% \thickmuskip=0mu
% \nulldelimiterspace=0pt
% \scriptspace=0pt
\begin{split}
\mathcal{L}_{div} = \frac{\sum_{i=1}^{M}\Psi(S_i)}{M-1},
\end{split}
\end{equation}
where $M$, as before, is the number of source domains. Since each source classifier is paired with $M-1$ classifiers in the other domains, we normalize $\mathcal{L}_{div}$ by $M-1$.

In the next two sections we propose our method to quantify the function $\mathcal{I}$. As we stated before, to calculate $\mathcal{I}$, we need to calculate document transformation costs and the classifier capacity (or performance) in the feature space.

\subsection{Transformation Costs} \label{subsubsec:transform-cost}
To incorporate the magnitude of the transformation in our scoring criterion, we measure the distance between the representations of the source and the target documents when the model is trained only for the source task. The intuition is that the resulting distance is what we would try to eliminate if we trained the classifier for the target task. To implement this idea, we calculate the \textit{point-wise} distance between the following two covariance matrices: 1) the covariance matrix of the encoded source data  $E(\textsc{x}^{s})$--recall that $E$ is our domain encoder--when the classifier is fully trained on source data, and 2) the covariance matrix of the encoded target data $E(\textsc{x}^{t})$ using the same encoder.

We begin by calculating the covariance of the encoded source documents:
\begin{equation} \label{eq:source-covar}
\small
% \setlength{\jot}{0pt}
% \setlength{\abovedisplayskip}{5pt}
% \setlength{\belowdisplayskip}{5pt}
% \medmuskip=0mu
% \thinmuskip=0mu
% \thickmuskip=0mu
% \nulldelimiterspace=0pt
% \scriptspace=0pt
\begin{split}
C^s=\frac{\frac{1}{n_s} \sum_{i=1}^{n_s}~(E(\textsc{x}_{i}^{s}) - \mu^s)^{\intercal} (E(\textsc{x}_{i}^{s}) - \mu^s)}{1-\frac{1}{n_s}},
\end{split}
\end{equation}
where $(\parm)^{\intercal}$ is the matrix transpose, $n_s$ is the number of source documents, and $\mu^s$ is the mean of the source representations: $\mu^s=\frac{\sum_{i}^{n_s}~E(\textsc{x}_{i}^{s})}{n_s}$. Then we calculate the point-wise covariance (i.e., the contribution of one document to the covariance matrix) of the target data:
\begin{equation} \label{eq:target-covar}
\small
% \setlength{\jot}{0pt}
% \setlength{\abovedisplayskip}{5pt}
% \setlength{\belowdisplayskip}{5pt}
% \medmuskip=0mu
% \thinmuskip=0mu
% \thickmuskip=0mu
% \nulldelimiterspace=0pt
% \scriptspace=0pt
\begin{split}
C_{r}^{t}=\frac{\frac{1}{n_t} (E(\textsc{x}_{r}^{t}) - \mu^t)^{\intercal} (E(\textsc{x}_{r}^{t}) - \mu^t)}{1-\frac{1}{n_t}},
\end{split}
\end{equation}
where $\textsc{x}_{r}^{t}$ is the \textit{r-th} target document, $n_t$ is the number of target documents, and $\mu^t$ is the mean of the target representations. Equations \ref{eq:source-covar} and \ref{eq:target-covar} are the weighted formulation of covariance matrix, see \citet{covar-extensions} for the theoretical derivation of these two equations. Finally, the transformation cost $d(S,\textsc{x}_{r}^{t})$ for the \textit{r-th} document in the target domain with respect to the source domain $S$ is achieved by measuring the difference between the two covariance matrices:
\begin{equation} \label{eq:transform-cost}
\small
% \setlength{\jot}{0pt}
% \setlength{\abovedisplayskip}{5pt}
% \setlength{\belowdisplayskip}{5pt}
% \medmuskip=0mu
% \thinmuskip=0mu
% \thickmuskip=0mu
% \nulldelimiterspace=0pt
% \scriptspace=0pt
\begin{split}
d(S,\textsc{x}_{r}^{t})=\left\Vert C_{r}^{t} - C^s \right\Vert_{F}^{2},
\end{split}
\end{equation}
where $\left\Vert \parm \right\Vert_{F}$ is the matrix Frobenius norm. 

% Intuitively, we measure the contribution of the document $\textsc{x}_{r}^{t}$ to the covariance matrix of target data (Equation \ref{eq:target-covar}), then we compare the resulting matrix with the covariance of source representations (Equation \ref{eq:transform-cost}) to obtain the magnitude of the divergence.

\subsection{Classifier Capacities in Target Domain} \label{subsubsec:classifier-capacity}
The prediction error of a classifier in the target domain is measurable only if we have access to labeled target documents. However, this is not the case in unsupervised Domain Adaptation. Therefore, we aim to estimate this quantity via a proxy task. Even though target labels are not available, the density of target data points after the transformation is available. Note that in the regions that the ratio of source data points to target data points is high, the classification error rate in the target domain is probably low. Thus, we use this ratio as an approximation to the classification performance. To calculate this density ratio we use the LogReg model \cite{logreg,transfer-cabibr} which derives the ratio from the output of a logistic regression classifier:
\begin{equation} \label{eq:classify-capacity}
\small
% \setlength{\jot}{0pt}
% \setlength{\abovedisplayskip}{5pt}
% \setlength{\belowdisplayskip}{5pt}
% \medmuskip=0mu
% \thinmuskip=0mu
% \thickmuskip=0mu
% \nulldelimiterspace=0pt
% \scriptspace=0pt
\begin{split}
q(S,\textsc{x})=&\frac{P_S(\textsc{x})}{P_T(\textsc{x})}=\frac{P(T)P(S|\textsc{x})}{P(S)P(T|\textsc{x})},
\end{split}
\end{equation}
where $S$ is the source domain, $\textsc{x}$ is a sample document, $q$ is the desired ratio for the document $\textsc{x}$, $P_S(\textsc{x})$ and $P_T(\textsc{x})$ are the distribution of source and target documents respectively; and $P(T)$ and $P(S)$ are the probabilities of picking a random document from the target and the source domains. These values are constant, and can be calculated directly, i.e., $P(T)=\frac{n_t}{n_t+n_s}$. The quantities $P(S|\textsc{x})$ and $P(T|\textsc{x})$ are the probabilities of classifying the document $\textsc{x}$ as source or target respectively. These values can be calculated by training a logistic regression classifier on the transformed data and probabilistically labeling documents as source or target.

To calculate our indicator function $\mathcal{I}$, defined in Section \ref{subsec:pairing}, we first rank the source domains based on their deemed reliability for classifying each target document, thereafter, we select the domain with the highest reliability score. In order to disproportionally relate this score to the transformation cost (i.e., $d(S,\textsc{x}_{r}^{t})$) and to proportionally relate it to the classifier capacity (i.e., $q(S,\textsc{x}_{r}^{t})$) we use a product equation as follows:
\begin{equation} \label{eq:pair-score}
\small
% \setlength{\jot}{0pt}
% \setlength{\abovedisplayskip}{5pt}
% \setlength{\belowdisplayskip}{5pt}
% \medmuskip=0mu
% \thinmuskip=0mu
% \thickmuskip=0mu
% \nulldelimiterspace=0pt
% \scriptspace=0pt
\begin{split}
Score(S,\textsc{x}_{r}^{t})=q(S,\textsc{x}_{r}^{t}) \times e^{d(S,\textsc{x}_{r}^{t})^{-1}},
\end{split}
\end{equation}
where $Score(S,\textsc{x}_{r}^{t})$ is the reliability score of the source domain $S$ with respect to the target document $\textsc{x}_{r}^{t}$. In the experiments we observed that if we directly use the inverse of $d(\parm)$ in the product, due to the magnitude of this term, the resulting value dominates the entire score. We empirically observed that an exponential inverse term yields in better results. Given this reliability score, the most reliable source domain is selected by the function $\mathcal{I}$, i.e.,:
\begin{equation} \label{eq:pair-indicator}
\small
% \setlength{\jot}{0pt}
% \setlength{\abovedisplayskip}{3pt}
% \setlength{\belowdisplayskip}{3pt}
% \medmuskip=0mu
% \thinmuskip=0mu
% \thickmuskip=0mu
% \nulldelimiterspace=0pt
% \scriptspace=0pt
\begin{split}
\mathcal{I}(S_i,\textsc{x}_{r}^{t})=\left\{\begin{matrix}
1 & i=\operatorname*{arg\,max}_k~~Score(S_k,\textsc{x}_{r}^{t}) \\
0 & otherwise.
\end{matrix}\right.
\end{split}
\end{equation}
Therefore, given the source domain $S_i$, this function returns 1 if the target document $\textsc{x}_{r}^{t}$ has a low transformation cost and also with a high probability is correctly labeled by the classifier in $S_i$, otherwise it returns 0. 

%% add a paragraph on the paired classifier here (see the code, it is "shared kl")
In Section \ref{subsec:pairing}, we proposed a method for pairing the source classifiers. While the aim of this method is primarily to help the under-performing source classifiers, it is also a medium for transferring knowledge across the document encoders. In this regard, this method is not fully efficient because the transfer of knowledge in this method is unidirectional--from the reliable classifiers to the others. In order to fully share the knowledge obtained in one encoder with all the other encoders, we add one medium classifier to every encoder. To train the medium classifiers, we pair their outputs with all the source classifiers using a KL-divergence loss term. The medium classifiers are not used for testing, they are discarded after the training. We denote the objective term for training the medium classifiers by $\mathcal{L}_{med}$. In the next section, we describe our training procedure.

\subsection{Training} \label{subsec:train}
To train our model, we first carry out the algorithm proposed in Section \ref{subsec:coordinate} to create the coordination between encoders and also to obtain the pseudo-optimal values of $\lambda$ in each source domain. To pair source classifiers (Section \ref{subsec:pairing}), we need the transformation costs (Section \ref{subsubsec:transform-cost}) and the classifier capacities (Section \ref{subsubsec:classifier-capacity}). To calculate the transformation costs, we train each source classifier once and use Equation \ref{eq:transform-cost} to obtain the costs. To calculate the classifier capacities we need the vector representations of the encoded source and target documents. Thus, while executing the algorithm in Section \ref{subsec:coordinate}, we also store the document vectors to calculate the classifier capacities using Equation \ref{eq:classify-capacity}. Finally, the entire model is trained using the loss function below:
\begin{equation} \label{eq:all-loss}
\small
% \setlength{\jot}{0pt}
% \setlength{\abovedisplayskip}{0pt}
% \setlength{\belowdisplayskip}{0pt}
% \medmuskip=0mu
% \thinmuskip=0mu
% \thickmuskip=0mu
% \nulldelimiterspace=0pt
% \scriptspace=0pt
\begin{split}
\mathcal{L}= \sum_{i}^{M} &\{ J(C_i(\textsc{x}^{s_i}), y^{s_i}) ~ + \lambda_i \mathcal{D}(E_i(\textsc{x}^{s_i}), E_i(\textsc{x}^{t})) \} \\ 
& + \alpha \mathcal{L}_{div} + \mathcal{L}_{med},
\end{split}
\end{equation}
where $M$ is the number of source domains, $C_i$ is the \textit{i-th} source classifier, $J$ is the cross-entropy function, $\lambda_i$ is the \textit{i-th} pseudo-optimal value of the scale factors, $E_i$ is the \textit{i-th} domain encoder, $\mathcal{L}_{div}$ is the divergence loss described in Section \ref{subsec:pairing}, and $\alpha$ is a penalty term. The penalty term prevents the classifiers from converging to one single classifier by governing the magnitude of the divergence loss. In order to further preserve the diversity between source classifiers during the training we gradually decrease the penalty term $[\alpha \to 0]$. $\mathcal{L}_{med}$ is the loss term to train the medium classifiers. $\mathcal{D}$ is the discrepancy objective between source and target representations; for this term we used the correlation alignment loss introduced by \citet{coral}. 

% In this section we described our approach to multiple-source unsupervised Domain Adaptation. Our model can be summarized in two steps: 1)~coordinating domain encoders using model hyper-parameters estimated by a proxy task, 2)~pairing source classifiers using the transformation costs and the classification errors. For each step we provided an intuition, and then, formulated our solution with enough details such that one can reproduce our results.

% \noindent\textbf{A note on time complexity of our model.} Our model is trained efficiently. The algorithm described in Section \ref{subsec:coordinate} can be viewed as the regular grid search. In fact, this algorithm is potentially faster than the grid search in Domain Adaptation, because we directly use the target domain rather than the set of source domains as meta-targets.

%%%%%%%%%%%%%%%%%%%%%%%%%%%%%%%%%%%%%%%%%%%%%%%%%%%%%%%%%%%%%%%%%%%%%%%%
\section{Experimental Setup} \label{sec:setup}

% However, The following items can be found in the appendix with a more comprehensive detail: 1) more information about the datasets, 2) detailed information about the baselines and a short description about each one, 3) detailed information about the training and the tuning procedures.

% In this section we provide an overview of our experimental setup. Due to the space constraint we discuss the primary topics only. An extended revision of this article will be available on arXiv.org.

We evaluate our model in two datasets: 1) a dataset on detecting illness reports\footnote{Available at \url{https://github.com/p-karisani/illness-dataset}} with 4 domains \cite{illness-dt}, and 2) another dataset on detecting the reports of incidents with 4 domains that we compiled from the data collected by \citet{crisis-standard-dt}. Thus, the entire data consists of 8 domains. Each domain contains anything between 5,100 to 6,400 documents--the total of 44,984 documents. Apart from this data, we compiled a third smaller dataset on reporting disease cases for tuning the hyper-parameters of the baselines. The tuning dataset primarily contains the data collected by \citet{validation-dt}.\footnote{Our code, along the Incident and the tuning datasets that we compiled, will be available at: \url{https://github.com/p-karisani/CEPC} }

We compare our model with the following baselines: DAN \cite{dan}, CORAL \cite{coral}, JAN \cite{jan}, M3SDA \cite{moment-match}, and MDDA \cite{dom-ada-distil}. In the cases of DAN, CORAL, and JAN we compare with both the best source domain and the combined source domains--indicated by the suffixes ``-b'' and ``-c'' respectively. Additionally, we compare with the best and combined source-only models, i.e., training a classifier on the source and evaluating on the target. This amounts to the total of 10 baselines.

The classifiers based on pretrained transformers are state-of-the-art \cite{self-pretraining}. We used the pretrained bert-base model \cite{bert} as the domain encoder in our model and all of the baselines. As the classifier, we used a network with one tanh hidden layer of size 768 followed by a binary softmax layer. This domain encoder and classifier were used in all baselines. We carried out all of the experiments 5 times and report the average.\footnote{The algorithm in Section \ref{subsec:coordinate} is similar to a grid search. There is no need to retake this step in each iteration.  Thus, we ran this step for 5 times with different random seeds. Then we took the average of the results and cached to be used in all of our experiments.} Because the datasets are imbalanced the models were tuned for the F1 measure and to construct the document mini-batches in source domains we used a balanced sampler with replacement--the batch-size was 50 in all experiments. During the tuning we observed that the higher number of training epochs cause overfitting on the source domains, therefore, we fixed the number of training epochs to~3. Our model has one hyper-parameter--the penalty term $\alpha$ in the final objective function. The optimal quantity of this hyper-parameter in \VALIDATIONDT dataset is 0.9. We set the range of the scale factors $\lambda$ to $\{1.0, 0.1, 0.01, 0.001, 0.0001\}$ in our algorithm in Section \ref{subsec:coordinate}.

\section{Results and Analysis} \label{sec:result}

\noindent\textbf{Results.} Table \ref{tbl:result-detail} reports the results across the domains, and Table \ref{tbl:result-average} reports the average results in \INCIDENTDT and \ILLNESSDT datasets--we also included the in-domain results (labeled \textit{ORACLE}) by training the classifier on 80\% of each domain and evaluating on the remainder. We observe a substantial performance deterioration by comparing \textit{ORACLE} and \textit{Source-b}, which signifies the necessity of Domain Adaptation. Additionally, we observe that in some cases \textit{Source-b} outperforms \textit{Source-c}, which indicates that simply aggregating the source domains is not an effective strategy. The results in Table \ref{tbl:result-detail} testify that compared to the baseline models, \METHOD is the top performing model in the majority of the cases. Table \ref{tbl:result-average} reports the average results, again we see that on average our model is state-of-the-art.

One particularly interesting observation from Table \ref{tbl:result-detail} is that in certain cases \textit{Source-b} or \textit{Source-c} outperform all of the models (e.g., in Earthquake or Alzheimer's). This demonstrates the efficacy of pretraining under domain shift, which calls for further investigation into this area.
\begin{table*}
\centering
\small
\begin{tabu}{p{0.15in} p{0.55in} p{0.55in} p{0.55in} p{0.65in} p{0.55in} p{0.55in} p{0.55in} p{0.65in}} \Xhline{2\arrayrulewidth}
 \cline{1-9} & \multicolumn{4}{c}{\textbf{F1 in \INCIDENTDT dataset}}  &
 \multicolumn{4}{c}{\textbf{F1 in \ILLNESSDT dataset}} \\
\cmidrule[\heavyrulewidth](l){2-5} \cmidrule[\heavyrulewidth](l){6-9}
\multicolumn{1}{c}{\textbf{Method}} & 
\textbf{Explosion} & \textbf{Flood} & \textbf{Hurricane} & \textbf{Earthquake} 
& \textbf{Cancer} & \textbf{Diabetes} & \textbf{Parkinson's} & \textbf{Alzheimer's} 
\\ \Xhline{3\arrayrulewidth}
\multicolumn{1}{c}{\textit{ORACLE}} & 0.968$\pm$0.01 & 0.902$\pm$0.01 & 0.749$\pm$0.01 & 0.619$\pm$0.01 
& 0.865$\pm$0.00 & 0.828$\pm$0.02 & 0.857$\pm$0.02 & 0.792$\pm$0.04 \\
\multicolumn{1}{c}{\textit{Source-b}} & 0.739$\pm$0.03 & 0.794$\pm$0.01 & 0.674$\pm$0.02 & \textbf{0.505$\pm$0.00} 
& 0.771$\pm$0.00 & 0.769$\pm$0.00 & 0.816$\pm$0.00 & 0.799$\pm$0.00 \\
\multicolumn{1}{c}{\textit{Source-c}} & 0.734$\pm$0.02 & 0.777$\pm$0.02 & 0.617$\pm$0.01 & 0.489$\pm$0.01 
& 0.801$\pm$0.00 & 0.746$\pm$0.03 & 0.822$\pm$0.01 & \textbf{0.802$\pm$0.00} \\ \Xhline{3\arrayrulewidth} 
\multicolumn{1}{c}{\textit{DAN-b}} & 0.799$\pm$0.02 & 0.835$\pm$0.00 & 0.640$\pm$0.01 & 0.479$\pm$0.00 
& 0.772$\pm$0.01 & 0.779$\pm$0.00 & 0.781$\pm$0.01 & 0.775$\pm$0.00 \\ 
\multicolumn{1}{c}{\textit{CORAL-b}} & 0.760$\pm$0.02 & 0.823$\pm$0.00 & 0.678$\pm$0.01 & 0.501$\pm$0.00 
& 0.773$\pm$0.00 & 0.764$\pm$0.01 & 0.816$\pm$0.00 & 0.789$\pm$0.00 \\ 
\multicolumn{1}{c}{\textit{JAN-b}} & 0.733$\pm$0.01 & 0.812$\pm$0.00 & 0.682$\pm$0.01 & \textbf{0.505$\pm$0.00} 
& 0.772$\pm$0.00 & 0.758$\pm$0.01 & 0.825$\pm$0.00 & 0.786$\pm$0.00 \\  \hline

\multicolumn{1}{c}{\textit{DAN-c}} & 0.857$\pm$0.01 & 0.840$\pm$0.00 & 0.665$\pm$0.01 & 0.450$\pm$0.01 
& 0.762$\pm$0.01 & 0.789$\pm$0.00 & 0.792$\pm$0.01 & 0.740$\pm$0.02 \\ 
\multicolumn{1}{c}{\textit{CORAL-c}} & 0.850$\pm$0.01 & 0.841$\pm$0.01 & 0.685$\pm$0.01 & 0.453$\pm$0.02 
& 0.779$\pm$0.01 & 0.786$\pm$0.00 & 0.820$\pm$0.00 & 0.772$\pm$0.00 \\ 
\multicolumn{1}{c}{\textit{JAN-c}} & 0.836$\pm$0.02 & 0.842$\pm$0.00 & 0.680$\pm$0.02 & 0.451$\pm$0.01 
& 0.777$\pm$0.01 & 0.785$\pm$0.01 & 0.818$\pm$0.00 & 0.762$\pm$0.01 \\ 
\multicolumn{1}{c}{\textit{MDDA}} & 0.722$\pm$0.07 & 0.822$\pm$0.02 & 0.643$\pm$0.03 & 0.431$\pm$0.01 
& 0.746$\pm$0.02 & 0.767$\pm$0.03 & 0.794$\pm$0.02 & 0.758$\pm$0.01 \\
\multicolumn{1}{c}{\textit{M3SDA}} & 0.802$\pm$0.02 & 0.849$\pm$0.00 & 0.667$\pm$0.01 & 0.455$\pm$0.01 
& 0.743$\pm$0.00 & 0.783$\pm$0.00 & 0.765$\pm$0.01 & 0.713$\pm$0.03 \\  \hline 
\multicolumn{1}{c}{\textit{\METHOD}} & \textbf{0.905$\pm$0.01} & \textbf{0.857$\pm$0.01} & \textbf{0.739$\pm$0.01} & 0.468$\pm$0.01 
& \textbf{0.808$\pm$0.01} & \textbf{0.795$\pm$0.00} & \textbf{0.848$\pm$0.00} & 0.794$\pm$0.00 \\ \Xhline{3\arrayrulewidth}
\end{tabu}
\caption{F1 in the datasets. The baselines show mixed results. \METHOD is the most consistent one.} \label{tbl:result-detail}
\vspace{-0.2cm}
\end{table*}
\begin{table}
\centering
\small
\begin{tabu}{p{0.15in} p{0.24in} p{0.24in} p{0.3in}  p{0.24in} p{0.24in} p{0.24in} } \Xhline{2\arrayrulewidth}
 \cline{1-7} & \multicolumn{3}{c}{\textbf{\INCIDENTDT dataset}}  &
 \multicolumn{3}{c}{\textbf{\ILLNESSDT dataset}} \\
\cmidrule[\heavyrulewidth](l){2-4} \cmidrule[\heavyrulewidth](l){5-7}
\multicolumn{1}{c}{\textbf{Method}} & 
\textbf{F1} & \textbf{Pre.} & \textbf{Rec.} & 
\textbf{F1} & \textbf{Pre.} & \textbf{Rec.} \\ \Xhline{3\arrayrulewidth}
\multicolumn{1}{c}{\textit{ORACLE}} & 0.809 & 0.775 & 0.849 & 0.835 & 0.811 & 0.861 \\
\multicolumn{1}{c}{\textit{Source-b}} & 0.678 & \textbf{0.738} & 0.642 & 0.789 & 0.762 & 0.820 \\
\multicolumn{1}{c}{\textit{Source-c}} & 0.654 & 0.735 & 0.643 & 0.793 & \textbf{0.791} & 0.807 \\ \Xhline{3\arrayrulewidth}
\multicolumn{1}{c}{\textit{DAN-b}} & 0.688 & 0.683 & 0.737 & 0.777 & 0.720 & 0.852 \\ 
\multicolumn{1}{c}{\textit{CORAL-b}} & 0.691 & 0.701 & 0.710 & 0.786 & 0.740 & 0.845 \\ 
\multicolumn{1}{c}{\textit{JAN-b}} & 0.683 & 0.705 & 0.687 & 0.785 & 0.745 & 0.838 \\  \hline

\multicolumn{1}{c}{\textit{DAN-c}} & 0.703 & 0.659 & 0.828 & 0.771 & 0.658 & \textbf{0.935} \\  
\multicolumn{1}{c}{\textit{CORAL-c}} & 0.707 & 0.681 & 0.804 & 0.790 & 0.709 & 0.899 \\ 
\multicolumn{1}{c}{\textit{JAN-c}} & 0.702 & 0.682 & 0.793 & 0.785 & 0.702 & 0.901 \\ 
\multicolumn{1}{c}{\textit{MDDA}} & 0.654 & 0.571 & \textbf{0.847} & 0.766 & 0.659 & 0.921 \\ 
\multicolumn{1}{c}{\textit{M3SDA}} & 0.693 & 0.660 & 0.814 & 0.751 & 0.639 & 0.925 \\  \hline 
\multicolumn{1}{c}{\textit{\METHOD}} & \textbf{0.742} & 0.725 & 0.809 & \textbf{0.811} & 0.786 & 0.841 \\ \Xhline{3\arrayrulewidth}
\end{tabu}
\caption{Average F1, Precision, and Recall in the datasets. \METHOD outperforms all the baselines.} \label{tbl:result-average}
\end{table}

\begin{figure}
    \centering
    \begin{subfigure}[t]{0.3\linewidth}
        \includegraphics[width=\linewidth]{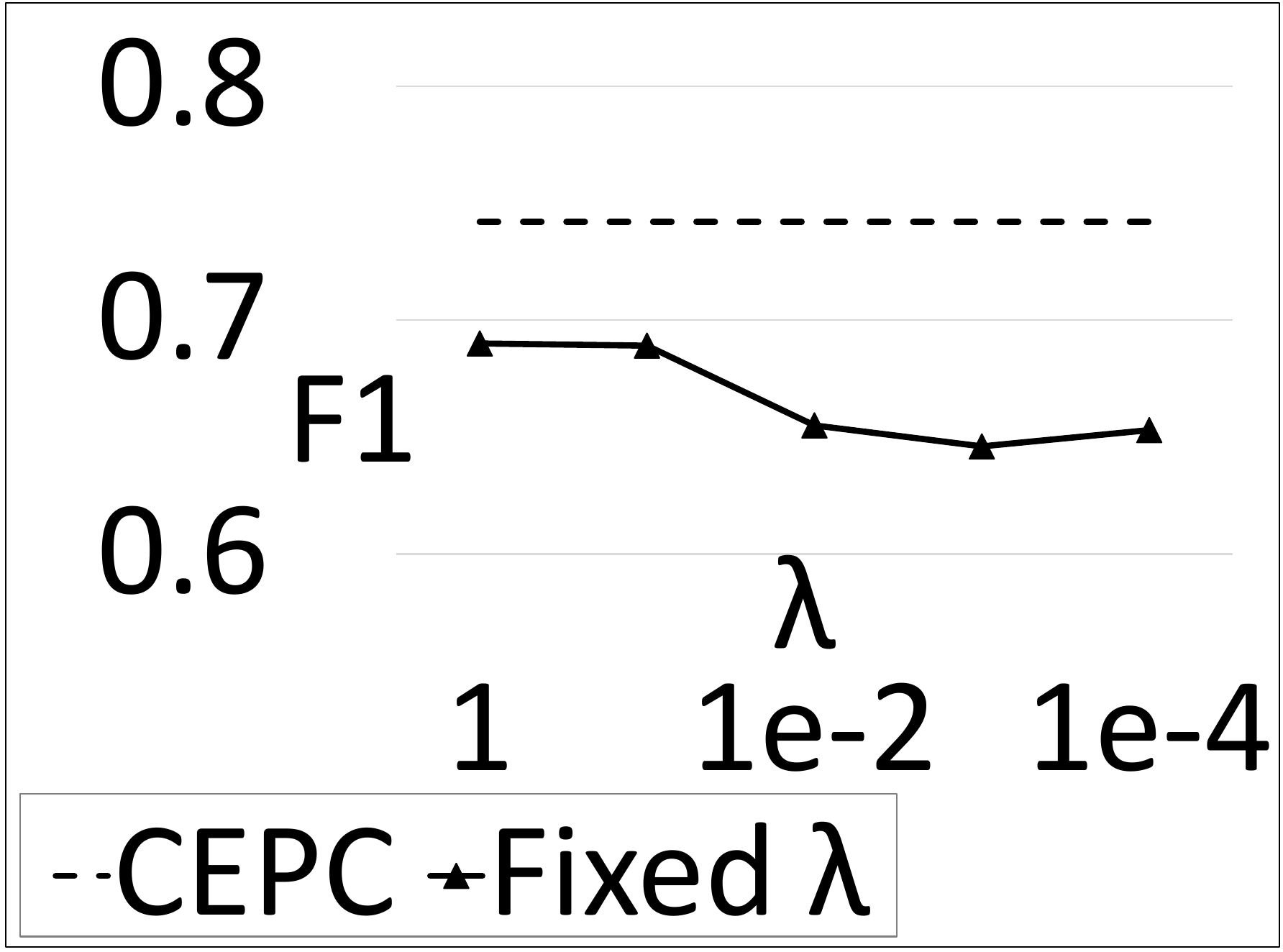}
        \caption{\scriptsize \METHOD vs. Fixed $\lambda$}
        \label{fig:lambda-curve-incident}
    \end{subfigure}\hfill
    \begin{subfigure}[t]{0.3\linewidth}
        \includegraphics[width=\linewidth]{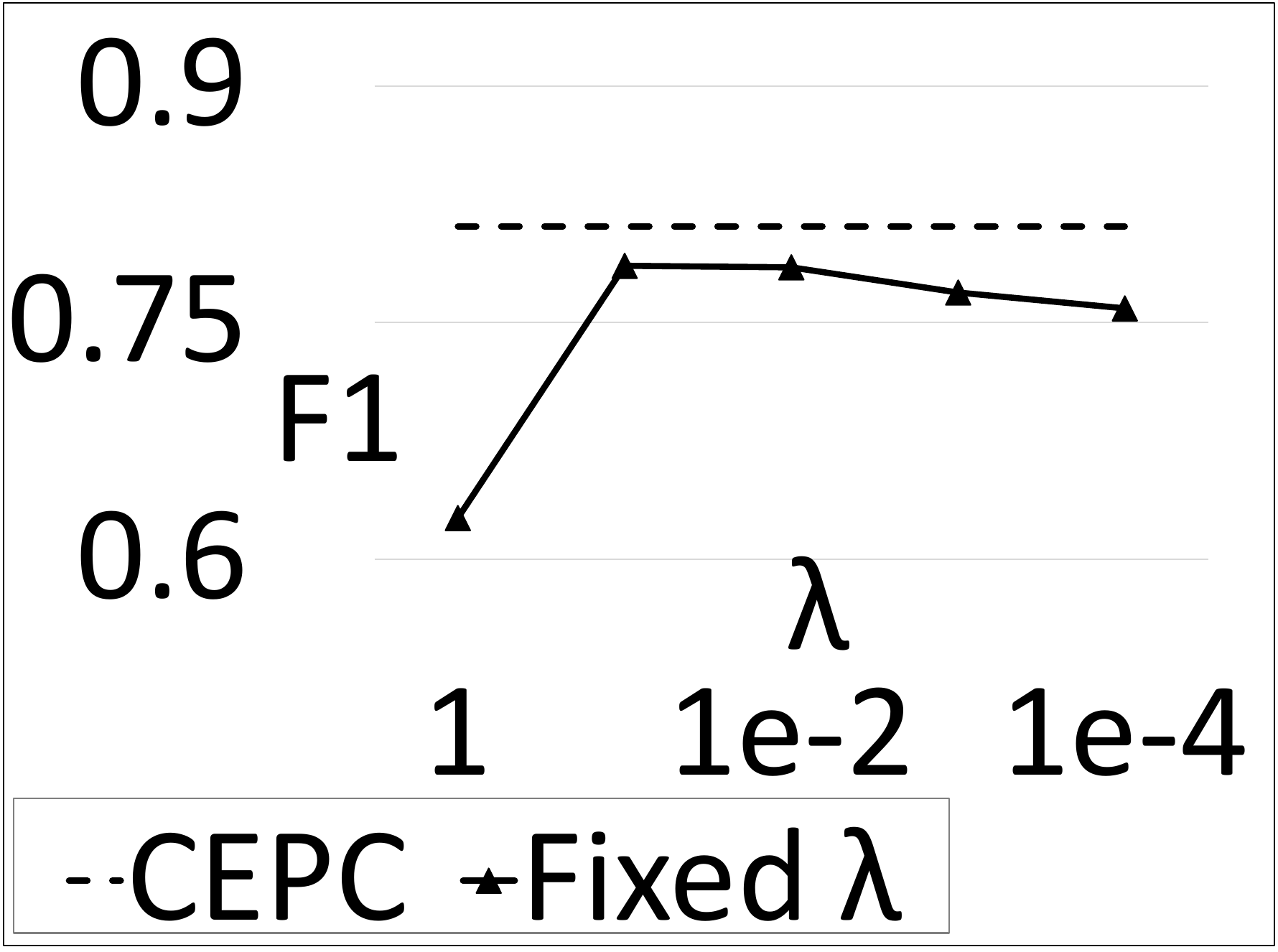}
        \caption{\scriptsize \METHOD vs. Fixed $\lambda$}
        \label{fig:lambda-curve-illness}
    \end{subfigure}\hfill
    \begin{subfigure}[t]{0.3\linewidth}
        \includegraphics[width=\linewidth]{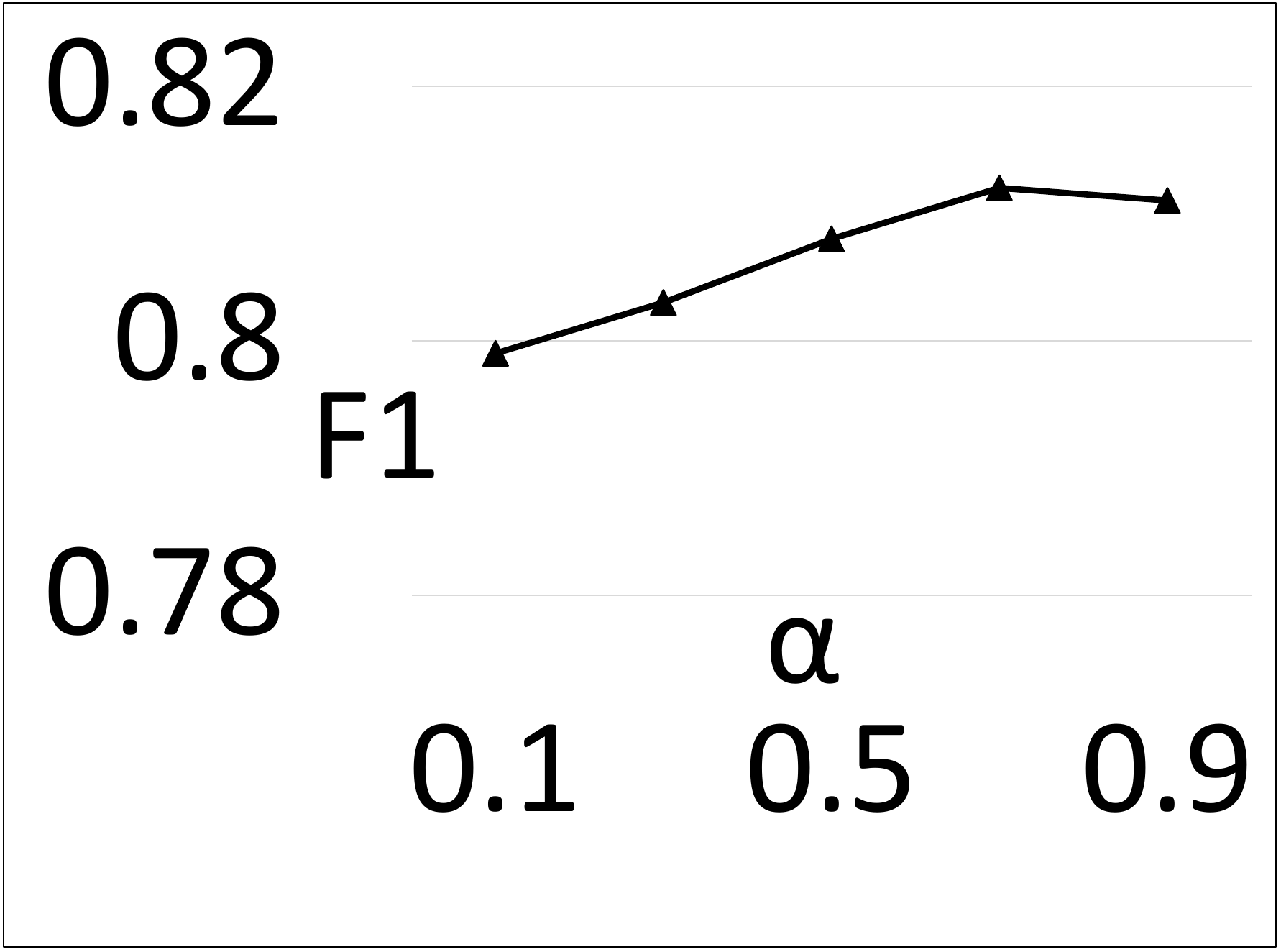}
        \caption{\scriptsize F1 vs. $\alpha$}
        \label{fig:alpha-curve-illness}
    \end{subfigure}
    \caption{\ref{fig:lambda-curve-incident}-\ref{fig:lambda-curve-illness}) \METHOD in comparison with a model with fixed scale factor $\lambda$ (Equation \ref{eq:all-loss}), in \INCIDENTDT (\ref{fig:lambda-curve-incident}) and \ILLNESSDT (\ref{fig:lambda-curve-illness}) datasets. \ref{fig:alpha-curve-illness}) Model sensitivity to the penalty term $\alpha$ (Equation \ref{eq:all-loss}) in \ILLNESSDT dataset.} \label{fig:lambda-curve}
\vspace{-0.2cm}
\end{figure}

\noindent\textbf{Empirical Analysis.} We begin by empirically demonstrating the efficacy of coordinating encoders (Section \ref{subsec:coordinate}). To this end, we compare \METHOD with a model that has a fixed scale factor $\lambda$.\footnote{We also gradually increased the value of $\lambda$ in each iteration, as proposed by \citet{gradient-reversal}.} Figures \ref{fig:lambda-curve-incident} and \ref{fig:lambda-curve-illness} report the results of this model at varying values of $\lambda$. The graphs show that even if we knew the optimal value of $\lambda$, \METHOD would still outperform this model\footnote{Note that, in reality the target labels are not available. Thus, we are comparing our model with the best case scenario.}--our model is shown by the dashed line. 

Next, we further investigate the utility of our algorithm for finding the pseudo-optimal values of hyper-parameters by comparing this algorithm with the traditional method of using source domains as meta-targets. Table \ref{tbl:meta-target} reports the results of a model that uses this technique. We see that again \METHOD outperforms this model. This suggests that the target domain possesses properties that cannot be recovered from the other source domains.
\begin{table}
\centering
\small
\begin{tabu}{p{0.7in} p{0.7in} p{0.3in} p{0.4in} p{0.4in} } \Xhline{3\arrayrulewidth}
\multicolumn{1}{c}{\textbf{Dataset}} & \textbf{Method} & \textbf{F1} & \textbf{Precision} & \textbf{Recall} \\ \Xhline{3\arrayrulewidth}
\multicolumn{1}{c}{\multirow{2}{*}{\INCIDENTDT}} & Meta-Target & 0.713 & 0.689 & 0.811 \\
\multicolumn{1}{c}{} & \METHOD & 0.742 & 0.725 & 0.809 \\ \hline
\multicolumn{1}{c}{\multirow{2}{*}{\ILLNESSDT}} & Meta-Target & 0.799 & 0.762 & 0.845 \\
\multicolumn{1}{c}{} & \METHOD & 0.811 & 0.786 & 0.841 \\ \Xhline{3\arrayrulewidth}
\end{tabu}
\caption{Using the source domains as meta-targets to tune for the scale factor $\lambda$ (Equation \ref{eq:all-loss}).} \label{tbl:meta-target}
\end{table}

\begingroup % this line and the next line reduce the table cell margins to fit in the page
\setlength{\tabcolsep}{3.5pt} 
\begin{table}
\centering
\small
\begin{tabu}{p{1.1in}  p{0.3in} p{0.5in} p{0.4in} } \Xhline{3\arrayrulewidth}
\multicolumn{1}{c}{\textbf{Method}} & \textbf{F1} & \textbf{Precision} & \textbf{Recall} \\ \Xhline{3\arrayrulewidth}
\multicolumn{1}{c}{\textit{\METHOD}} & 0.811 & 0.786 & 0.841 \\ \Xhline{3\arrayrulewidth}
\multicolumn{1}{c}{\METHOD w/o \textit{Paired Classifier}} & 0.793 & 0.713 & 0.895 \\ 
\multicolumn{1}{c}{\METHOD w/o \textit{Classifier Capacity}} & 0.801 & 0.782 & 0.826 \\ 
\multicolumn{1}{c}{\METHOD w/o \textit{Transformation Cost}} & 0.809 & 0.783 & 0.839 \\ 
\Xhline{3\arrayrulewidth}
\end{tabu}
\caption{Ablation study of the paired classifiers (Sections \ref{subsec:pairing}, \ref{subsubsec:transform-cost}, and \ref{subsubsec:classifier-capacity}) in \ILLNESSDT dataset.} \label{tbl:pairing-ablation}
\end{table}
\endgroup % this line resets the table cell margins to default

We also demonstrate whether pairing source classifiers (Section \ref{subsec:pairing}) is effective. Additionally, we show the extent in which the transformation costs (Section \ref{subsubsec:transform-cost}) and the classifier capacities (Section \ref{subsubsec:classifier-capacity}) contribute to the performance. To this end, we report an ablation study on these modules in Table \ref{tbl:pairing-ablation}. We see that this technique is indeed effective, and also each one of its underlying steps are contributing.

Finally, we report the sensitivity of \METHOD to the hyper-parameter $\alpha$ in Figure \ref{fig:alpha-curve-illness} (Equation \ref{eq:all-loss}). We see that as the value of $\alpha$ increases, the performance slightly improves. 

% The graph indicates that there exists an optimal point for the degree of pairing, in which after that the performance deteriorates. This indicates that larger values of $\alpha$ cause a convergence between source classifiers.

In this work we demonstrated that several domain adaptation models that previously have shown to be competitive, not necessarily perform well in user-generated data, specifically in social media data. Our experiments suggest that every area may have particular properties and call for further investigation into under-explored areas. Future work may evaluate the efficacy of existing models in languages other than English, specifically those that have lower resources and are richer in morphology, e.g., Persian and Kurdish.

% There are several exciting directions to explore as future work. First, given the large margin between \textit{ORACLE} and the best models in Table \ref{tbl:result-average}, a long way is still ahead. 

% In summary, in this section we empirically analyzed every single module of our model. We also evaluated our model in multiple settings and showed that it outperforms the state of the art in the majority of tasks, including those with pretrained transformers. The results in Tables \ref{tbl:result-detail} and \ref{tbl:result-average} clearly demonstrate that our study is pushing the state of the art in this area. However, given the large margin between \textit{ORACLE} and the best models in Table \ref{tbl:result-average}, a long way is still ahead. 

% One particular direction that we may explore is investigating the role of pretraining in overcoming the challenges of learning under domain shift. Table~\ref{tbl:result-detail} shows that in certain cases--i.e., Earthquake and Alzheimer's domains--none of the exisiting domain adaptation models can outperform a pretrained source-only classifier. These models explicitly align the distribution of source and target documents. We believe it is worthwhile to explore pretraining tasks specifically geared to enhance learning under domain shift.

%%%%%%%%%%%%%%%%%%%%%%%%%%%%%%%%%%%%%%%%%%%%%%%%%%%%%%%%%%%%%%%%%%%%%%%%
\section{Conclusions} \label{sec:conclusion}

In this work, we proposed a novel model for multiple-source unsupervised Domain Adaptation. Our model possesses two essential properties: 1) It can automatically coordinate domain encoders by inferring their update rates via a proxy task. 2) It can pair source classifiers based on their predicated error rates in the target domain. We carried out experiments in two datasets and showed that our model outperforms the state of the art.

\section*{Acknowledgments}
We thank the anonymous reviewers for their insightful feedback.

% \begin{small}
\bibliography{aaai22.bib}

\begin{thebibliography}{37}
\providecommand{\natexlab}[1]{#1}

\bibitem[{Alam et~al.(2020)Alam, Sajjad, Imran, and Ofli}]{crisis-standard-dt}
Alam, F.; Sajjad, H.; Imran, M.; and Ofli, F. 2020.
\newblock Standardizing and Benchmarking Crisis-related Social Media Datasets
  for Humanitarian Information Processing.
\newblock \emph{arXiv preprint arXiv:2004.06774}.

\bibitem[{Ben-David et~al.(2010)Ben-David, Blitzer, Crammer, Kulesza, Pereira,
  and Vaughan}]{dom-ada-theory}
Ben-David, S.; Blitzer, J.; Crammer, K.; Kulesza, A.; Pereira, F.; and Vaughan,
  J.~W. 2010.
\newblock A theory of learning from different domains.
\newblock \emph{Machine learning}, 79(1-2): 151--175.

\bibitem[{Ben-David and Schuller(2003)}]{dom-ada-first}
Ben-David, S.; and Schuller, R. 2003.
\newblock Exploiting Task Relatedness for Multiple Task Learning.
\newblock In Sch{\"o}lkopf, B.; and Warmuth, M.~K., eds., \emph{Learning Theory
  and Kernel Machines}, 567--580. Berlin, Heidelberg: Springer Berlin
  Heidelberg.

\bibitem[{Biddle et~al.(2020)Biddle, Joshi, Liu, Paris, and Xu}]{validation-dt}
Biddle, R.; Joshi, A.; Liu, S.; Paris, C.; and Xu, G. 2020.
\newblock \emph{Leveraging Sentiment Distributions to Distinguish Figurative
  From Literal Health Reports on Twitter}, 1217–1227.
\newblock New York, NY, USA: Association for Computing Machinery.
\newblock ISBN 9781450370233.

\bibitem[{Caruana(1997)}]{multi-task}
Caruana, R. 1997.
\newblock Multitask Learning.
\newblock \emph{Mach. Learn.}, 28(1): 41--75.

\bibitem[{Cui and Bollegala(2020)}]{dom-ada-multi-attention}
Cui, X.; and Bollegala, D. 2020.
\newblock Multi-source Attention for Unsupervised Domain Adaptation.
\newblock \emph{arXiv preprint arXiv:2004.06608}.

\bibitem[{Devlin et~al.(2019)Devlin, Chang, Lee, and Toutanova}]{bert}
Devlin, J.; Chang, M.-W.; Lee, K.; and Toutanova, K. 2019.
\newblock {BERT}: Pre-training of Deep Bidirectional Transformers for Language
  Understanding.
\newblock In \emph{Proc of the 2019 NAACL}, 4171--4186.

\bibitem[{Ganin and Lempitsky(2015)}]{gradient-reversal}
Ganin, Y.; and Lempitsky, V.~S. 2015.
\newblock Unsupervised Domain Adaptation by Backpropagation.
\newblock In Bach, F.~R.; and Blei, D.~M., eds., \emph{Proceedings of the 32nd
  International Conference on Machine Learning, {ICML} 2015, Lille, France,
  6-11 July 2015}, volume~37 of \emph{{JMLR} Workshop and Conference
  Proceedings}, 1180--1189. JMLR.org.

\bibitem[{Guo, Pasunuru, and Bansal(2020)}]{da-bandit}
Guo, H.; Pasunuru, R.; and Bansal, M. 2020.
\newblock Multi-Source Domain Adaptation for Text Classification via
  DistanceNet-Bandits.
\newblock In \emph{The Thirty-Fourth {AAAI} Conference on Artificial
  Intelligence, {AAAI} 2020, The Thirty-Second Innovative Applications of
  Artificial Intelligence Conference, {IAAI} 2020, The Tenth {AAAI} Symposium
  on Educational Advances in Artificial Intelligence, {EAAI} 2020, New York,
  NY, USA, February 7-12, 2020}, 7830--7838. {AAAI} Press.

\bibitem[{Guo, Shah, and Barzilay(2018)}]{da-mixture}
Guo, J.; Shah, D.~J.; and Barzilay, R. 2018.
\newblock Multi-Source Domain Adaptation with Mixture of Experts.
\newblock In Riloff, E.; Chiang, D.; Hockenmaier, J.; and Tsujii, J., eds.,
  \emph{Proceedings of the 2018 Conference on Empirical Methods in Natural
  Language Processing, Brussels, Belgium, October 31 - November 4, 2018},
  4694--4703. Association for Computational Linguistics.

\bibitem[{Haeusser et~al.(2017)Haeusser, Frerix, Mordvintsev, and
  Cremers}]{assoc-loss}
Haeusser, P.; Frerix, T.; Mordvintsev, A.; and Cremers, D. 2017.
\newblock Associative Domain Adaptation.
\newblock In \emph{Proceedings of the IEEE International Conference on Computer
  Vision (ICCV)}.

\bibitem[{Holzenberger, Blair-Stanek, and Van~Durme(2020)}]{legal-domain}
Holzenberger, N.; Blair-Stanek, A.; and Van~Durme, B. 2020.
\newblock A Dataset for Statutory Reasoning in Tax Law Entailment and Question
  Answering.
\newblock \emph{arXiv preprint arXiv:2005.05257}.

\bibitem[{Isobe et~al.(2021)Isobe, Jia, Chen, He, Shi, Liu, Lu, and Wang}]{ccl}
Isobe, T.; Jia, X.; Chen, S.; He, J.; Shi, Y.; Liu, J.; Lu, H.; and Wang, S.
  2021.
\newblock Multi-Target Domain Adaptation With Collaborative Consistency
  Learning.
\newblock In \emph{{IEEE} Conference on Computer Vision and Pattern
  Recognition, {CVPR} 2021, virtual, June 19-25, 2021}, 8187--8196. Computer
  Vision Foundation / {IEEE}.

\bibitem[{Karisani and Karisani(2020)}]{our-corona}
Karisani, N.; and Karisani, P. 2020.
\newblock Mining coronavirus (covid-19) posts in social media.
\newblock \emph{arXiv preprint arXiv:2004.06778}.

\bibitem[{Karisani and Agichtein(2018)}]{wespad}
Karisani, P.; and Agichtein, E. 2018.
\newblock Did You Just Have a Heart Attack?: Towards Robust Detection of
  Personal Health Mentions in Social Media.
\newblock In \emph{Proc of the 2018 WWW}, 137--146.
\newblock ISBN 978-1-4503-5639-8.

\bibitem[{Karisani, Agichtein, and Ho(2020)}]{co-decomp}
Karisani, P.; Agichtein, E.; and Ho, J. 2020.
\newblock Domain-Guided Task Decomposition with Self-Training for Detecting
  Personal Events in Social Media.
\newblock In \emph{Proceedings of The Web Conference 2020}, WWW ’20,
  2411–2420. New York, NY, USA: Association for Computing Machinery.

\bibitem[{Karisani, Choi, and Xiong(2021)}]{view-distill}
Karisani, P.; Choi, J.~D.; and Xiong, L. 2021.
\newblock View Distillation with Unlabeled Data for Extracting Adverse Drug
  Effects from User-Generated Data.
\newblock arXiv:2105.11354.

\bibitem[{Karisani and Karisani(2021)}]{self-pretraining}
Karisani, P.; and Karisani, N. 2021.
\newblock Semi-Supervised Text Classification via Self-Pretraining.
\newblock In \emph{Proceedings of the 14th ACM International Conference on Web
  Search and Data Mining}, WSDM '21, 40–48. Association for Computing
  Machinery.

\bibitem[{Karisani, Karisani, and Xiong(2021)}]{illness-dt}
Karisani, P.; Karisani, N.; and Xiong, L. 2021.
\newblock Contextual Multi-View Query Learning for Short Text Classification in
  User-Generated Data.
\newblock arXiv:2112.02611.

\bibitem[{Long et~al.(2015)Long, Cao, Wang, and Jordan}]{dan}
Long, M.; Cao, Y.; Wang, J.; and Jordan, M.~I. 2015.
\newblock Learning Transferable Features with Deep Adaptation Networks.
\newblock In Bach, F.~R.; and Blei, D.~M., eds., \emph{Proceedings of the 32nd
  International Conference on Machine Learning, {ICML} 2015, Lille, France,
  6-11 July 2015}, volume~37 of \emph{{JMLR} Workshop and Conference
  Proceedings}, 97--105. JMLR.org.

\bibitem[{Long et~al.(2017)Long, Zhu, Wang, and Jordan}]{jan}
Long, M.; Zhu, H.; Wang, J.; and Jordan, M.~I. 2017.
\newblock Deep transfer learning with joint adaptation networks.
\newblock In \emph{International conference on machine learning}, 2208--2217.
  PMLR.

\bibitem[{Mansour, Mohri, and Rostamizadeh(2009)}]{dom-ada-multi}
Mansour, Y.; Mohri, M.; and Rostamizadeh, A. 2009.
\newblock Domain adaptation with multiple sources.
\newblock In \emph{Advances in neural information processing systems},
  1041--1048.

\bibitem[{Pan and Yang(2010)}]{transfer-survey}
Pan, S.~J.; and Yang, Q. 2010.
\newblock A Survey on Transfer Learning.
\newblock \emph{{IEEE} Trans. Knowl. Data Eng.}, 22(10): 1345--1359.

\bibitem[{Peng et~al.(2019)Peng, Bai, Xia, Huang, Saenko, and
  Wang}]{moment-match}
Peng, X.; Bai, Q.; Xia, X.; Huang, Z.; Saenko, K.; and Wang, B. 2019.
\newblock Moment Matching for Multi-Source Domain Adaptation.
\newblock In \emph{2019 {IEEE/CVF} International Conference on Computer Vision,
  {ICCV} 2019, Seoul, Korea (South), October 27 - November 2, 2019},
  1406--1415. {IEEE}.

\bibitem[{Price(1972)}]{covar-extensions}
Price, G.~R. 1972.
\newblock Extension of covariance selection mathematics.
\newblock \emph{Annals of human genetics}, 35(4): 485--490.

\bibitem[{Qin(1998)}]{logreg}
Qin, J. 1998.
\newblock Inferences for case-control and semiparametric two-sample density
  ratio models.
\newblock \emph{Biometrika}, 85(3): 619--630.

\bibitem[{Sun and Saenko(2016)}]{coral}
Sun, B.; and Saenko, K. 2016.
\newblock Deep {CORAL:} Correlation Alignment for Deep Domain Adaptation.
\newblock In Hua, G.; and J{\'{e}}gou, H., eds., \emph{Computer Vision - {ECCV}
  2016 Workshops - Amsterdam, The Netherlands, October 8-10 and 15-16, 2016,
  Proceedings, Part {III}}, volume 9915 of \emph{Lecture Notes in Computer
  Science}, 443--450.

\bibitem[{Torralba and Efros(2011)}]{domain-shift}
Torralba, A.; and Efros, A.~A. 2011.
\newblock Unbiased look at dataset bias.
\newblock In \emph{CVPR 2011}, 1521--1528. IEEE.

\bibitem[{Tzeng et~al.(2017)Tzeng, Hoffman, Saenko, and Darrell}]{adda}
Tzeng, E.; Hoffman, J.; Saenko, K.; and Darrell, T. 2017.
\newblock Adversarial Discriminative Domain Adaptation.
\newblock In \emph{2017 {IEEE} Conference on Computer Vision and Pattern
  Recognition, {CVPR} 2017, Honolulu, HI, USA, July 21-26, 2017}, 2962--2971.
  {IEEE} Computer Society.

\bibitem[{Wang et~al.(2020)Wang, Long, Wang, and Jordan}]{transfer-cabibr}
Wang, X.; Long, M.; Wang, J.; and Jordan, M.~I. 2020.
\newblock Transferable Calibration with Lower Bias and Variance in Domain
  Adaptation.
\newblock \emph{CoRR}, abs/2007.08259.

\bibitem[{Wright and Augenstein(2020)}]{dom-ada-mixture-bert}
Wright, D.; and Augenstein, I. 2020.
\newblock Transformer Based Multi-Source Domain Adaptation.
\newblock \emph{arXiv preprint arXiv:2009.07806}.

\bibitem[{Xu et~al.(2018)Xu, Chen, Zuo, Yan, and Lin}]{cocktail}
Xu, R.; Chen, Z.; Zuo, W.; Yan, J.; and Lin, L. 2018.
\newblock Deep cocktail network: Multi-source unsupervised domain adaptation
  with category shift.
\newblock In \emph{Proceedings of the IEEE Conference on Computer Vision and
  Pattern Recognition}, 3964--3973.

\bibitem[{Yang et~al.(2020{\natexlab{a}})Yang, Balaji, Lim, and
  Shrivastava}]{dom-ada-curriculum-2}
Yang, L.; Balaji, Y.; Lim, S.-N.; and Shrivastava, A. 2020{\natexlab{a}}.
\newblock Curriculum Manager for Source Selection in Multi-Source Domain
  Adaptation.
\newblock \emph{arXiv preprint arXiv:2007.01261}.

\bibitem[{Yang et~al.(2020{\natexlab{b}})Yang, Wang, Gao, Shrivastava,
  Weinberger, Chao, and Lim}]{semi-sup-dom-ada-cotrain}
Yang, L.; Wang, Y.; Gao, M.; Shrivastava, A.; Weinberger, K.~Q.; Chao, W.-L.;
  and Lim, S.-N. 2020{\natexlab{b}}.
\newblock MiCo: Mixup Co-Training for Semi-Supervised Domain Adaptation.
\newblock \emph{arXiv preprint arXiv:2007.12684}.

\bibitem[{Yosinski et~al.(2014)Yosinski, Clune, Bengio, and
  Lipson}]{how-transferable}
Yosinski, J.; Clune, J.; Bengio, Y.; and Lipson, H. 2014.
\newblock How transferable are features in deep neural networks?
\newblock In \emph{Advances in neural information processing systems},
  3320--3328.

\bibitem[{Zellinger et~al.(2019)Zellinger, Moser, Grubinger, Lughofer,
  Natschläger, and Saminger-Platz}]{central-loss}
Zellinger, W.; Moser, B.~A.; Grubinger, T.; Lughofer, E.; Natschläger, T.; and
  Saminger-Platz, S. 2019.
\newblock Robust unsupervised domain adaptation for neural networks via moment
  alignment.
\newblock \emph{Information Sciences}, 483: 174 -- 191.

\bibitem[{Zhao et~al.(2020)Zhao, Wang, Zhang, Gu, Li, Song, Xu, Hu, Chai, and
  Keutzer}]{dom-ada-distil}
Zhao, S.; Wang, G.; Zhang, S.; Gu, Y.; Li, Y.; Song, Z.; Xu, P.; Hu, R.; Chai,
  H.; and Keutzer, K. 2020.
\newblock Multi-source distilling domain adaptation.
\newblock In \emph{Proceedings of the AAAI Conference on Artificial
  Intelligence}, volume~34, 12975--12983.

\end{thebibliography}
% \end{small}

\end{document}